\documentclass[conference,review,anonymous]{IEEEtran}
\IEEEoverridecommandlockouts
\usepackage{multirow}
\usepackage{cite}
\usepackage{amsmath,amssymb,amsfonts}
\usepackage{algorithmic}
\usepackage{graphicx}
\usepackage{textcomp}
\usepackage{xcolor}
\usepackage{booktabs}
\usepackage{lmodern}
\usepackage{enumitem}
\usepackage{caption}
\usepackage{graphicx}
\usepackage{subfigure}

\def\BibTeX{{\rm B\kern-.05em{\sc i\kern-.025em b}\kern-.08em
    T\kern-.1667em\lower.7ex\hbox{E}\kern-.125emX}}

\begin{document}

\title{Supervised Contrastive Vision Transformer for Breast Histopathological Image Classification}

\author{\IEEEauthorblockN{Mohammad Shiri}
\IEEEauthorblockA{\textit{Department of Computer Science} \\
\textit{Old Dominion University}\\
Norfolk, VA, USA \\
mshir001@odu.edu}
\and
\IEEEauthorblockN{Monalika Padma Reddy}
\IEEEauthorblockA{\textit{Department of Computer Science} \\
\textit{Old Dominion University}\\
Norfolk, VA, USA \\
mpadm001@odu.edu}
\and
\IEEEauthorblockN{Jiangwen Sun}
\IEEEauthorblockA{\textit{Department of Computer Science} \\
\textit{Old Dominion University}\\
Norfolk, VA, USA \\
jsun@odu.edu}
}

\maketitle

\begin{abstract}

Invasive ductal carcinoma (IDC) is the most prevalent form of breast cancer. Breast tissue histopathological examination is critical in diagnosing and classifying breast cancer. Although existing methods have shown promising results, there is still room for improvement in the classification accuracy and generalization of IDC using histopathology images. We present a novel approach, Supervised Contrastive Vision Transformer (SupCon-ViT), for improving the classification of invasive ductal carcinoma in terms of accuracy and generalization by leveraging the inherent strengths and advantages of both transfer learning, i.e., pre-trained vision transformer, and supervised contrastive learning. Our results on a benchmark breast cancer dataset demonstrate that SupCon-Vit achieves state-of-the-art performance in IDC classification, with an F1-score of 0.8188, precision of 0.7692, and specificity of 0.8971, outperforming existing methods. In addition, the proposed model demonstrates resilience in scenarios with minimal labeled data, making it highly efficient in real-world clinical settings where labelled data is limited. Our findings suggest that supervised contrastive learning in conjunction with pre-trained vision transformers appears to be a viable strategy for an accurate classification of IDC, thus paving the way for a more efficient and reliable diagnosis of breast cancer through histopathological image analysis.

\end{abstract}


\section{Introduction}

On a global scale, cancer continues to be a major fatality contributor. In the United States specifically, it is predicted that there will be a startling estimate of 1.9 million new cases of cancer in 2023, or an average of 5,370 cases every day. The anticipated numbers also show a grim prognosis of 609,820 people dying from cancer-related problems, averaging around 1,670 fatalities every day \cite{siegel2023cancer}. Breast cancer is the most common cancer that women are suffering from. It is an ailment that causes the breast cells to develop rapidly, leading to the formation of a lump in specific regions of the breast. Among the varying types of breast cancer, Invasive Ductal Carcinoma (IDC) is the one that is the most common, accounting for more than 80\% \cite{waxman1998scott} of all cases. 
Good prognosis of breast cancer requires early detection and screening. Mammography, breast MRI, and breast ultra-sonogram are the three methods used in breast cancer screening. 

Histopathological images \cite{gurcan2009histopathological} are extensively used in classifying and grading cancer. Histopathology slides provide more intricate details for diagnosis when compared to mammography, CT, and other imaging techniques. Analyzing histopathological images is a tedious and challenging process. Given the demand for professional understanding in this domain, the task of scanning vast tissue images necessitates the expertise of a pathologist. Consequently, a pathologist's experience may inadvertently have an adverse effect on the analysis results. 

The advances in WSI imaging technology, as well as an increasing quantity of published datasets \cite{janowczyk2016deep, mooney2017breasthistopathology}, have promoted the use of deep learning algorithms in digital histopathology. Several deep learning algorithms \cite{cruz2014automatic, janowczyk2016deep, shawi2022interpretable, araujo2017classification} and machine learning algorithms  \cite{amrane2018breast, michael2022optimized, allugunti2022breast, wang2016automatic} are used to detect IDC in histopathology images. Histopathologic imaging provides important visual information about tissue samples, and deep learning models have proven effective in the evaluation of histopathological images for cancer detection and diagnosis. Advances in computer vision have made automated breast cancer classification techniques increasingly popular. Deep learning techniques, in particular, have a great deal of potential in learning and extracting features from histopathology images that may go undetected in conventional laboratory testing. However, a significant amount of labeled data is required for training CNNs from scratch, which is a challenge here. 

One strategy to address this specific issue is transfer learning \cite{zhuang2020comprehensive}. Transfer learning allows the pre-trained model to be fine-tuned on a smaller task-specific dataset, potentially resulting in better performance with less data. Using histopathology images, IDC has been classified into positive (Malignant) and negative (benign) categories using CNN-based transfer learning models like DenseNet, ResNet, and VGG \cite{chaves2020evaluation, gonccalves2022cnn, farooq2020infrared, cabiouglu2020computer, roslidar2019study}. 

Although these methods have produced encouraging results, there is still room for improvement in the performance of IDC classification. In IDC diagnosis, contextual information and spatial relationships among several regions of interest within an image are crucial for an accurate classification. CNN models naturally capture local features, but they might not successfully include contextual data over the whole image. Contextual cues like the distribution of aberrant cells inside the tissue or relationships between various structures might offer crucial diagnostic information that CNN models could overlook. This may result in incorrect classification and erroneous diagnoses.  

Vision Transformer (ViT) \cite{dosovitskiy2020image} derives global representation from shallow levels, while shallow layers also provide local representation. Skip connections are significantly more powerful in ViT and have a significant influence on representation performance and similarity. A pre-trained ViT can be employed to capitalize on the model's ability to extract meaningful and significant features, resulting in improved performance on a specific task. Because the pre-trained ViT model already has an adequate understanding of visual features, this is especially valuable when there is insufficient labeled data, \cite{usman2022analyzing} thereby showing improved results when compared to training a ViT from scratch.

Utilizing contrastive learning \cite{chen2020simple} has recently resulted in a considerable improvement in self-supervised representation learning. A model is trained to separate the target image (also known as the "anchor") from several non-matching ("negative") images while simultaneously bringing the representations of an anchor and a matching ("positive") image together in embedding space. 

Supervised contrastive learning \cite{khosla2020supervised} bridges the gap between self-supervised and fully supervised learning by allowing contrastive learning to be used in a supervised setting.  It allows normalized embeddings of the same class to be pushed toward one another while embeddings of different classes are pushed away using labeled data. 
It allows for more positives per anchor. It generates a more diverse set of positive samples that nevertheless include semantically relevant information and label information that can be employed actively in the representation learning process as opposed to traditional contrastive learning. 

The primary objective of this study is to deliver more precise and dependable IDC classification results to medical practitioners, allowing them to make better patient treatment decisions. Our study aims to improve the accuracy of IDC classification using histopathological images by incorporating supervised contrastive learning with transfer learning i.e., pre-trained ViT. 

Our proposed approach involves fine-tuning the encoder of a pre-trained ViT model and incorporating supervised contrastive loss to classify IDC lesions into positive (Malignant) and negative (benign) categories. By leveraging the strengths of both techniques, our model can successfully learn and capture discriminative features for accurate IDC classification. The goal of integrating representation learning with spatial understanding is to improve the accuracy and reliability of IDC predictions. 

Our main contributions can be summarized as follows: \\ \\
1. We have introduced a novel approach called SupCon-ViT, that leverages supervised contrastive learning to fine-tune pre-trained vision transformers for binary classification on a breast cancer dataset. \\
2. Through our method, we demonstrate that the feature embedding generated is more distinctive, resulting in improved performance in predicting Malignant and Benign histopathology images. \\
3. We conducted experiments to identify the best hyperparameters and data augmentation techniques for training our models. \\
4. We compared the performance of transfer learning using our proposed method with existing CNN-based architectures, as well as simple Vision transformers, to establish the superiority of our work. \\



\section{Methodology}

\subsection{Dataset}

\begin{figure}[h!]
    \centering
    \subfigure[Benign]
    {\includegraphics[scale=0.7]{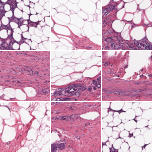}}
    \quad
    \subfigure[Malignant]
    {\includegraphics[scale=0.7]{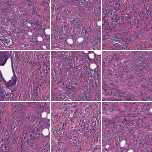}}
    \caption{The dataset comprises a total of 277,524 patches, each of 50x50 pixels, extracted from WSI. (a) A sample of images from a set of images that are benign, i.e., having a non-IDC (negative) diagnosis. (b) A sample of images from a set of images that are Malignant, i.e. have an IDC (positive) diagnosis.}
    \label{fig:Data}
\end{figure}

The dataset was curated here was curated by \cite{janowczyk2016deep} and was made publicly available on Kaggle \cite{mooney2017breasthistopathology}. It consists of digitized histopathology slides of breast cancer (BCa) from 279 women diagnosed with IDC at the Hospital of the University of Pennsylvania and The Cancer Institute of New Jersey. Each slide was digitized at 40x magnification and downscaled to a resolution of 4 µm/pixel due to their extremely large size. The ground truth IDC regions were obtained via manual delineation by an expert pathologist using ImageScope software. The dataset has 71.6\% images belonging to the malignant class and 18.4\% images belonging to the benign class. The annotations were made at 2x magnification, resulting in the inclusion of some stromal and non-invasive tissue. 

The dataset is split into three subsets, including 84 training and 29 validation cases for parameter exploration, and 166 test cases for final evaluation. Table \ref{tab:data}. shows the details of the train-test-validation split of the dataset. The dataset comprises 277,524 image patches, having 198,738 benign patches i.e. non-IDC (negative) IDC diagnosis, and 78,786 Malignant patches i.e. (positive) diagnoses. A total of 91,592 patches are used as train set, 28,677 patches are used as validation set. The remaining 157,255 patches are used as test set. Out of the 91,592 patches used in the training set, 65,351 patches are Benign, i.e. belong to non-IDC (negative) diagnosis and 26,241 patches are Malignant, i.e. belong to IDC (positive) diagnosis. 20401 patches in the validation are non-IDC (negative) diagnoses, i.e. benign and the remaining 8,276 patches are IDC (positive) diagnoses, i.e. Malignant. 112,986 patches in the test data are benign, i.e. non-IDC (negative) diagnosis, and the remaining 44,269 patches are malignant, i.e. belong to IDC (positive) diagnosis. Examples of positive (IDC) and negative (non-IDC) tissue regions from the dataset are provided in Figure \ref{fig:Data}.

\begin{table}[h]
\caption{The train, validation and test split of the 277,524 image patches in the IDC dataset}
\begin{center}
\begin{tabular}{lcccc}
 \toprule
 \textbf {Class} & \textbf{Train} & \textbf{Validation} & \textbf{Test}
\\
\midrule
Benign & 65351 & 20401 & 112986
\\
Malignant &  26241 & 8276 & 44269
\\
\bottomrule
\end{tabular}
\end{center}
\label{tab:data}
\end{table}

\subsection{SupCon-ViT}

The fundamental goal of this work is to improve the accuracy of IDC prediction. To do this, we leverage the powerful representation learning capabilities of the supervised contrastive loss. We adopt a two-step approach which also involves fine-tuning the encoder of the pretrained ViT model. We aim to further enhance the discriminant capacity of the learned feature representations by employing the supervised contrastive loss. This loss function encourages the model to push dissimilar IDC instances farther apart and to map similar IDC instances closer together in the feature space. 

\begin{figure*}[htp]
    \centering
    \includegraphics[scale=0.23] {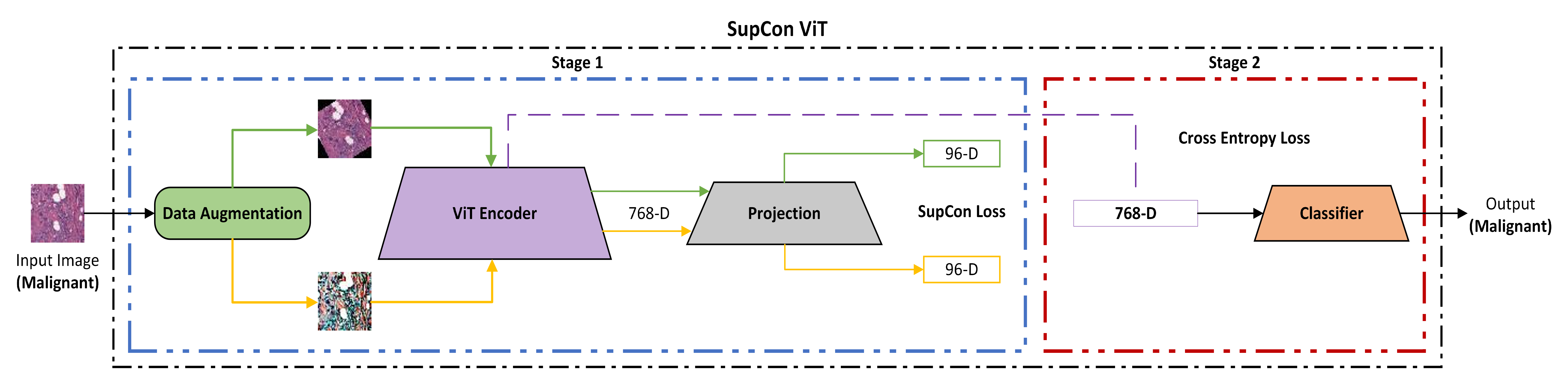}
    \caption{SupCon-ViT is trained in two stages - Stage 1 and Stage 2. Stage 1 comprises the data augmentation layer, the encoder(ViT) layer, and the projection layer. The supervised contrastive loss is calculated in Stage 1. Stage 2 has a classification layer which uses the cross-entropy loss and the frozen representations from Stage 1. }
    \setlength{\belowcaptionskip}{-10pt}
    \label{fig:SupConViT}
\end{figure*}

The accuracy is improved by effectively capturing and emphasizing the significant characteristics and patterns of IDC instances through model optimization based on this loss. Furthermore, we use the benefits of transfer learning in our method by employing a trained ViT model. Transfer learning allows us to leverage the ViT model's acquired knowledge and feature extraction skills. We exploit its rich visual representations and apply them to our target domain by fine-tuning this pre-trained model using our specific aim of predicting IDC. This technique improves the model's capacity to learn relevant IDC categorization characteristics and classify them accurately. The SupCon-ViT framework is depicted in Figure \ref{fig:SupConViT}.


The study adopts a training approach resembling Supervised Contrastive Learning \cite{khosla2020supervised}, consisting of two stages: pre-training and classifier training. Each stage includes multiple components and aspects aimed at optimizing the performance of the SupCon-ViT model. Three major components are used during the pre-training stage of the proposed framework; the data augmentation module, the encoder layer, and the projection layer. 

The training set comprises of a set of (image, label) pairs. In the data augmentation module, for each input image sample in the training set, two sets of augmentations are generated. Here, random vertical flip and random horizontal flip are used for data augmentation. Images are randomly flipped vertically along the horizontal axis via random vertical flips. Images are randomly flipped horizontally along the vertical axis with the help of random horizontal flip. Data augmentation strategies are critical for improving performance and generalizability. Using random vertical flips and random horizontal flips on the images helps increase the size and diversity of a training dataset. Variations that assist in increasing the robustness and generalization capacity of a model are introduced here. By employing the above mentioned data augmentation techniques, the proposed framework increases the diversity and variability of the training data, facilitating the learning of more robust and generalized IDC classification models.

The Encoder Network employed in the pre-training stage is from a pre-trained ViT model, where the transformer encoder is similar to the transformer encoder of \cite{dosovitskiy2020image}. The two sets of augmentations obtained from the Data Augmentation module are mapped into a 786 dimensional representation vector, r by the ViT encoder. This 768-dimensional representation vector, r, is forwarded through the projection network. The projection network is a single linear layer with a dimensionality of 96. This enables the initial 768-dimensional representation to be transformed into a more compact 96-dimensional space. The Projection Network outputs are l2 normalized to facilitate the measurement of distances within the projection space. The projection network outputs are used to compute the supervised contrastive loss, fostering the learning of discriminative feature embeddings. 

In order to more efficiently use label information, supervised contrastive loss pushes normalized embeddings from the same class to be closer together than embeddings from different classes. Mini-batches of data, which comprise both positive and negative samples, are used for computing this loss. The negative pairs are made up of augmented views from different examples in the batch, whereas the positive pairs are made up of two augmented views of the same example. The loss is subsequently calculated by computing the similarity between the positive and the negative pairs. It aims to minimize the discrepancy between the predicted output (embedding vectors) and the ground truth labels. It computes the loss by considering positive instances (patches) and their associated negative instances. The dot products between embedding vectors are scaled by the temperature parameter $\tau$ to control the importance of the similarity between vectors. The overall loss is the sum of these terms, with appropriate normalization based on the number of positive instances. \cite{khosla2020supervised} formulates the Supervised Contrastive Loss as:
\begin{equation*}
 L_{\text{SupCon}} = \sum_{i \in I} -\log\left(\frac{1}{|P(i)|} \sum_{p \in P(i)} \frac{\exp\left(z_i \cdot z_p / \tau\right)} {\sum_{a \in A(i)} \exp\left(z_i \cdot z_a / \tau\right)}\right)\end{equation*}

where $z_i$ is the representation of the anchor, with $z_p$ and $z_a$ representing the positive and arbitrary examples in the batch, respectively. The sets $P(i)$ and $A(i)$ refer to indices of samples sharing the same class as the anchor (excluding the anchor itself) and all indices in the batch, respectively. The parameter $\tau$ is a temperature scaling factor that adjusts the distribution's sharpness.

During the classifier training stage, the pre-training projection layer is discarded. A 786-dimensional representation vector generated during the pre-training phase from the encoder layer is frozen. The cross-entropy loss is subsequently used to train a linear classifier on this frozen representation vector.

\subsection{Model Training Setup and Hyper-parameters}

The Supervised Contrastive Loss, coupled with AdamW optimizer is used in the pre-training phase. The temperature parameter regulates the sharpness of the probability distribution in the contrastive loss computation and is set to 0.03. The learning rate determines the step size for updating the model parameters during optimization is set to $5\times10^{-5}$. 

The entire pre-training process runs through 50 epochs, with each epoch representing a complete iteration of an entire training dataset. The frozen feature representation acquired during the pre-training step is used in the classifier stage. A linear classifier is trained using the frozen features at a learning rate of 0.01, and the AdamW optimizer is used for parameter updates. An early stopping mechanism with a patience of 5, is used to minimize the training time. This implies that training is terminated if performance on the validation set does not improve for five successive epochs. 

Cross entropy Loss is used in the classifier stage to quantify the dissimilarity between predicted class probabilities and true labels. During the classifier training phase, the model is trained for a total of 50 epochs with the same early stopping mechanism. All the baseline models are trained on 64 mini batch sizes and in SupCon-ViT model in each iteration one mini batch of size 32 is selected from the dataset and two sets of its augmentation are fed to the network. 

\section{Results}

\subsection{Feature Extraction}

\begin{figure*}[!htp]
\centering
    {\includegraphics[scale=0.2]{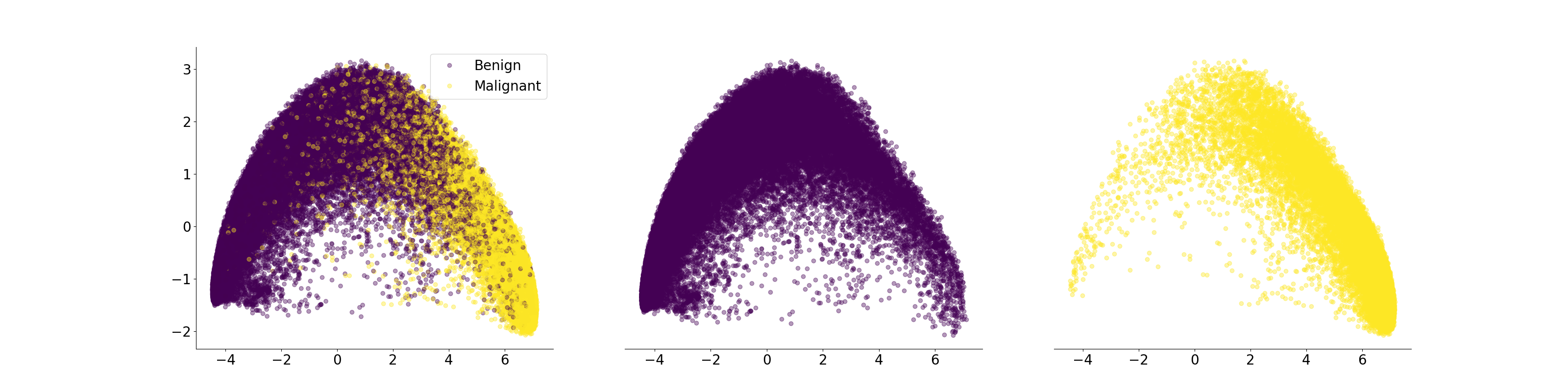}}
    \caption{PCA visualization of the feature representations of ViT on train set}
    \label{fig:TrainViT}
\end{figure*}

\begin{figure*}[!htp]
\centering
    {\includegraphics[scale=0.2]{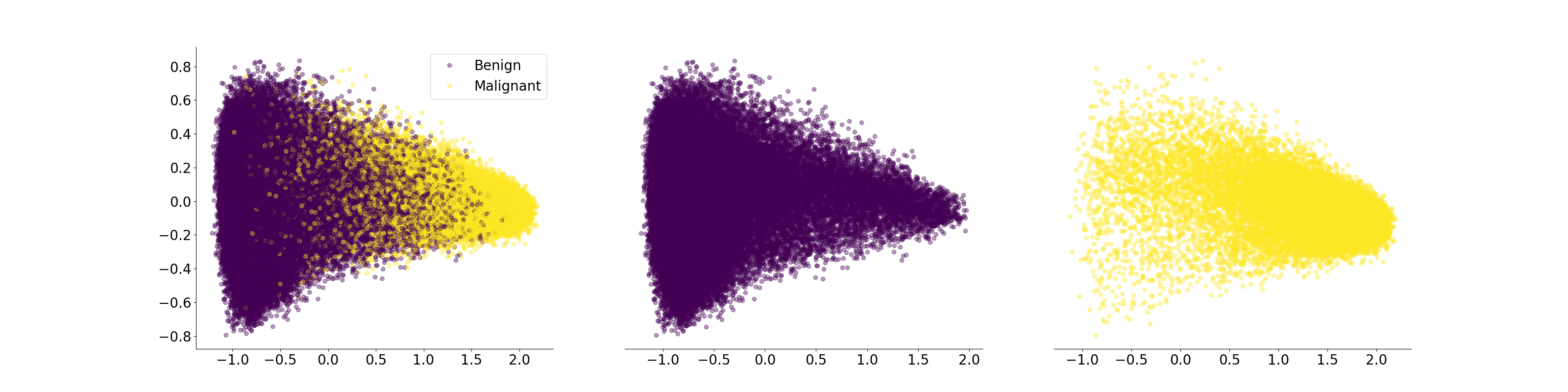}}
    \caption{PCA visualization of the feature representations of SupCon-ViT on train set}
    \label{fig:TrainCon}
\end{figure*}

We employed Principal Component Analysis (PCA) \cite{jolliffe1986principal}, a powerful visualization technique to gain insights into the feature embeddings of the SupCon-ViT model. We were able to discern similarities and differences in their learned representations by juxtaposing the visualizations derived from PCA for the SupCon-ViT and ViT. The results of these analyses shed light on how the models captured and structured the underlying features of the input data. 

Figure \ref{fig:TrainViT}. and Figure \ref{fig:TrainCon}. show the visualization of the feature embeddings of the SupCon-ViT and ViT on the training set. The purple color is used for representing the data points that are benign, i.e. IDC negative and the yellow color is used for representing the data points that are malignant, i.e. IDC positive. It can be seen from the feature embeddings, that SupCon-ViT exhibited a superior capacity for class separation in contrast to the embeddings generated by ViT. As demonstrated by the distinct separation and discernible clusters within its embeddings, the SupCon-ViT model revealed a substantial advantage in capturing and discriminating class-specific information. 

\subsection{Model Performance Evaluation}

We conducted a comprehensive comparison of the SupCon-ViT model's performance with other existing architectures, including ViT and a range of state-of-the-art pre-trained convolutional neural network (CNN) models. 

The chosen CNN architectures for comparison were AlexNet \cite{krizhevsky2017imagenet}, VGG16 \cite{simonyan2014very}, DenseNet121 \cite{huang2017densely}, ResNet50 \cite{he2016deep}, MobileNetV3Large \cite{howard2019searching}. After replacing the classification head, each of the models was fine-tuned utilizing the dataset with a train, test, and validation ratio as specified in Section 3.1 and a learning rate of $1\times10^{-5}$. 

\begin{figure}[htp]
    \includegraphics[scale=0.3] {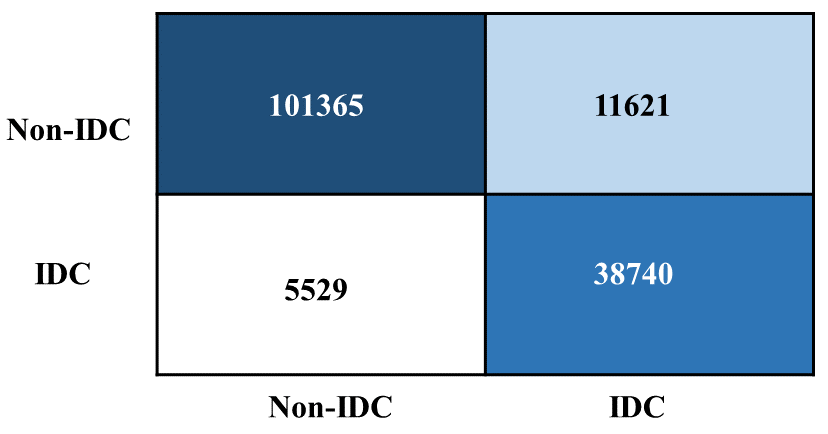}
    \caption{Confusion Matrix of the SupCon-ViT model on the test set}
    \setlength{\belowcaptionskip}{-10pt}
    \label{fig:ConfusionMat}
\end{figure}

We attempted to assess the relative strengths and shortcomings of the SupCon-ViT model in contrast to the other architectures by examining numerous metrics and performance indicators such as balanced accuracy, F1-score, Precision, and Recall. Table \ref{tab:Metric}. compares the proposed approach's performance to that of other architectures. The results indicate that the proposed model outperforms the other methods, with a higher F1-score of \textbf{0.8188}, balanced accuracy of \textbf{0.8861}, greater precision of \textbf{0.7692}, and enhanced specificity of \textbf{0.8971}. The confusion matrix for the model, which illustrates the model’s performance across different classes, is depicted in Fig \ref{fig:ConfusionMat}.

\begin{table*}[h]
\caption{Evaluating the performance of the proposed model, SupCon-ViT with ViT and other state-of-the-art CNN models on precision, specificity, F1-score, and balanced accuracy. SupCon-ViT model has better precision, specificity, F1-score, and balanced accuracy than the existing models.}
\begin{center}
\begin{tabular}{lcccccc} 
 \toprule
 \textbf{Model} & \textbf{Precision} & \textbf{Sensitivity} & \textbf{Specificity} & \textbf{F1-Score} & \textbf{ Balanced Accuracy} 
\\
 \midrule
\textbf{SupCon-ViT}&\textbf{0.7692} & 0.8751 & \textbf{0.8971} & \textbf{0.8188} & \textbf{0.8861}
\\
ViT & 0.6980 & \textbf{0.9256} & 0.8431 & 0.7959 & 0.8844 
\\
ResNet-50 & 0.6937 & 0.8775 & 0.8482 & 0.7749 & 0.8629 
\\
VGG-16 & 0.6893 & 0.8979 & 0.8414 & 0.7799 & 0.8697 
\\
DenseNet-121 & 0.7153 & 0.8750 & 0.8635 & 0.7871 & 0.8693
\\
MobileNet-V3Large & 0.7109 & 0.8790 & 0.8599 & 0.7861 & 0.8595 
\\
AlexNet & 0.6770 & 0.8768 & 0.8361 & 0.7641 & 0.8564
\\
\bottomrule
\end{tabular}
\end{center}
\label{tab:Metric}
\end{table*}

\subsection{IDC prediction maps}

\begin{figure*}[h!]
    \centering
    {{\includegraphics[scale=0.08]
    {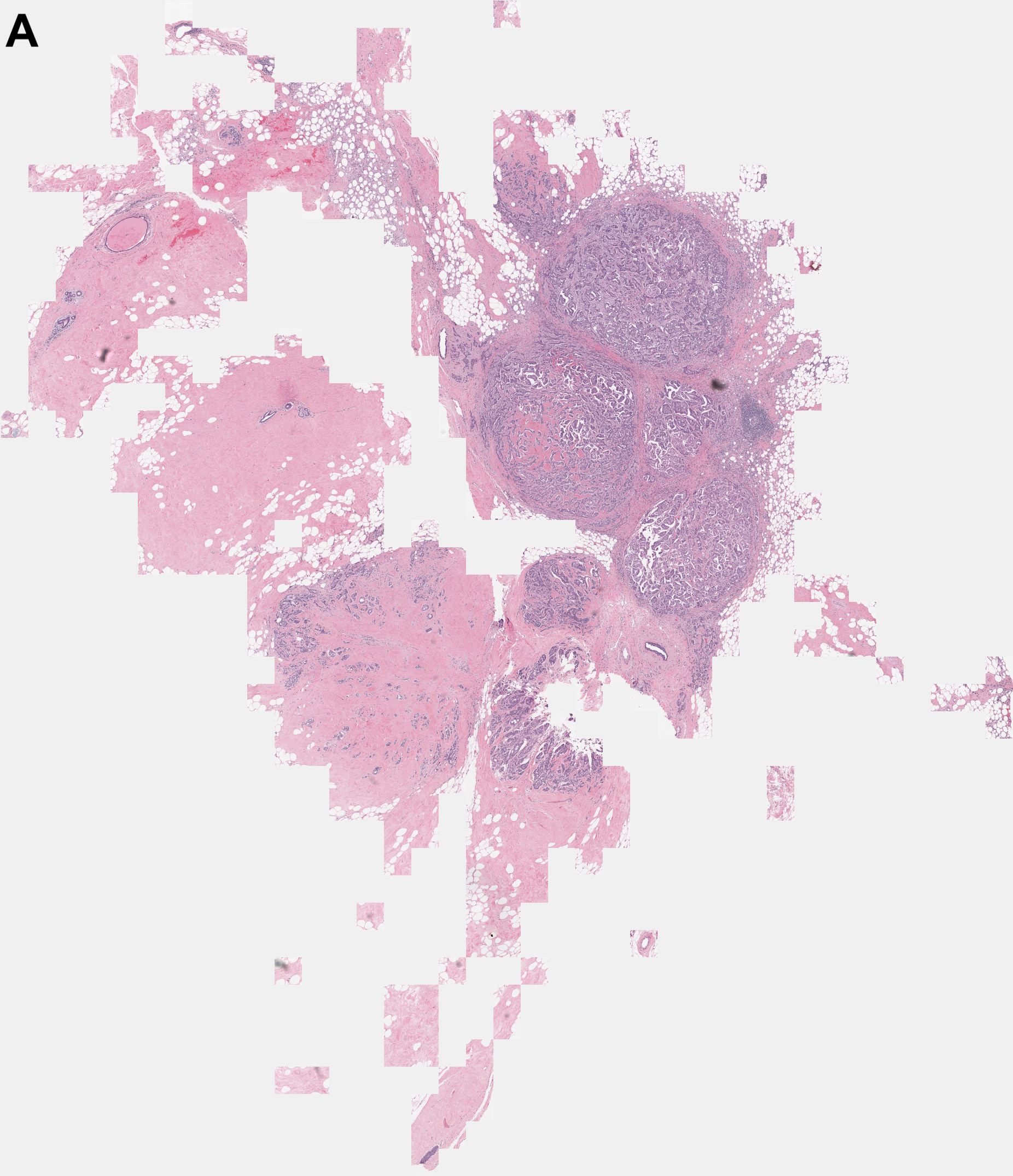}}}
    \quad
    {{\includegraphics[scale=0.08]{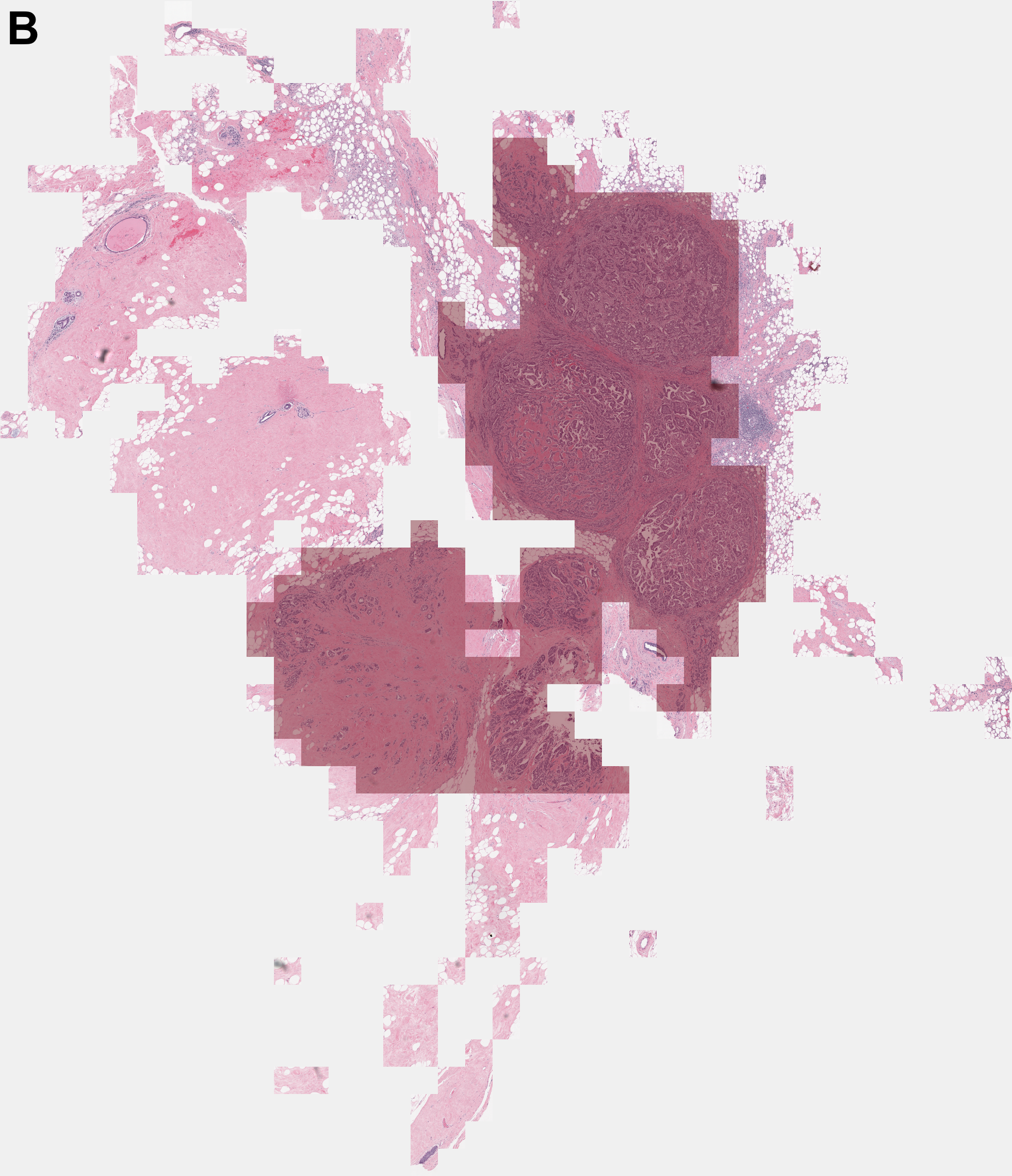}}}
    \quad
    {{\includegraphics[scale=0.08]{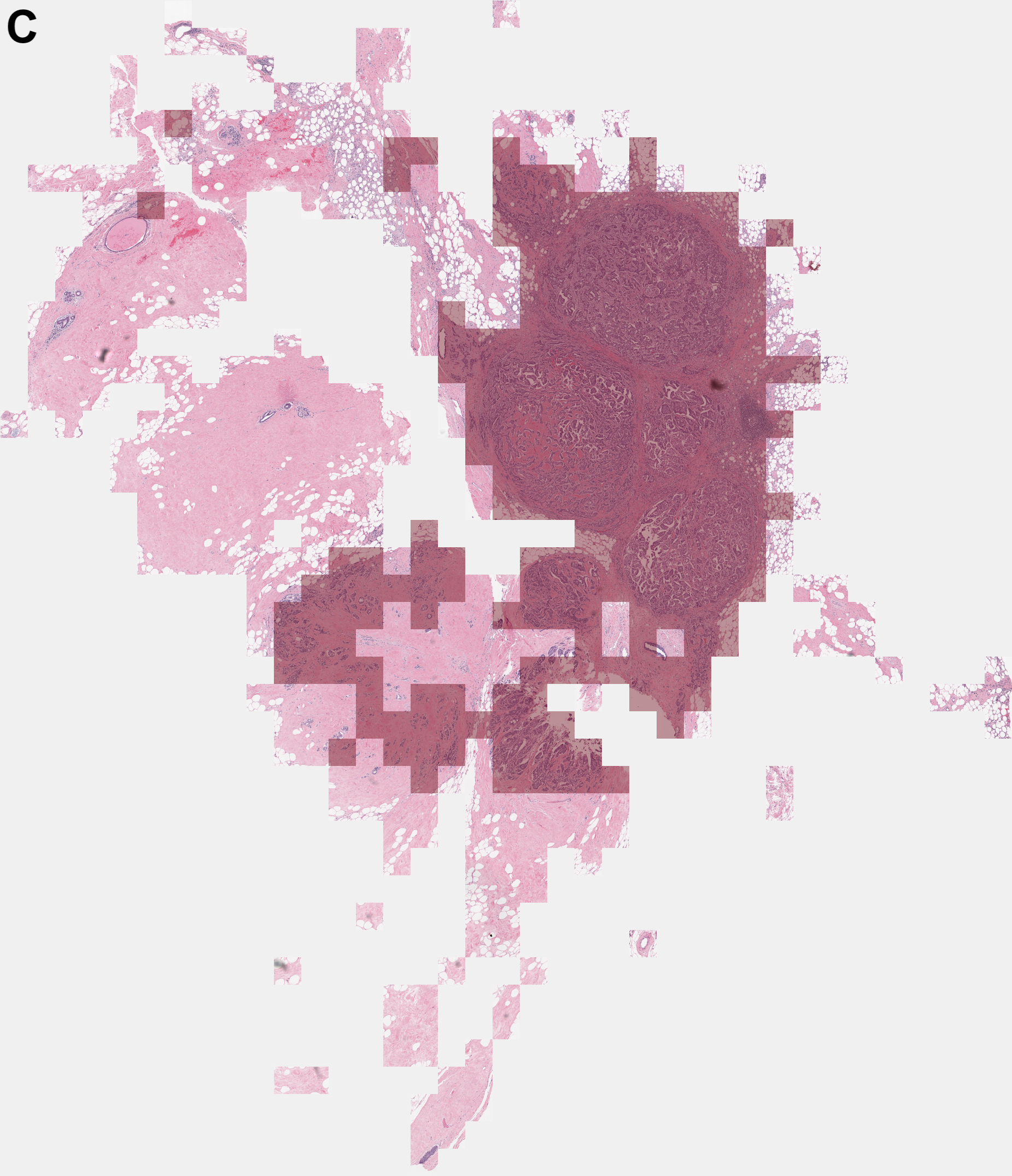}}}
    \caption{The prediction on a WSI image from one of the test subjects, obtained by the SupCon-ViT for Invasive Ductal Carcinoma(IDC). \textbf{A)} The whole slide image of a patient from the test set. \textbf{B)} The ground truth of the image of a patient from the test set. \textbf{C)} The predictions of the SupCon-ViT model on the image of a patient from the test set. The darker patches in both B and C indicate the Malignant regions i.e. IDC positive regions and the lighter patches indicate the benign regions i.e. IDC negative regions.}
    \label{fig:Heatmap}
\end{figure*}

We employed the SupCon-ViT model on the test set patches and obtained categorical values (0 for benign/non-IDC and 1 for malignant/IDC). Using these values, we reconstructed the whole slide image (WSI) for the patients in the test set. The SupCon-ViT efficiently distinguishes between IDC and non-IDC regions by leveraging the categorical values. This approach greatly enhances the computer-aided analysis of WSI samples, providing significant value in digital pathology research. Figure \ref{fig:Heatmap}. exhibits the prediction maps generated by the model for multiple subjects in the test set, demonstrating the model's ability to accurately identify IDC regions. Overall, the proposed model exhibits remarkable performance in identifying the IDC and non-IDC regions.

\subsection{Training Time Analysis}

The training times for each model were recorded using a single Nvidia V100 GPU. During the experiments, we trained the models, namely SupCon-ViT, DenseNet121, ViT, AlexNet, VGG16, MobileNetV3Large and ResNet50 on the IDC classification task. It is worth noting that the training times reported here are specific to our experimental setup and may vary depending on the hardware configuration and implementation details. 

From Table \ref{tab:Time}. the training times provide insights into the computational requirements of the models and demonstrate that the SupCon-ViT model has a relatively slower training time when compared to the other models. However, the convergence of training is quite fast and in our experiments with just one epoch, the model reaches the minima.

\begin{table}[h]
\caption{Comparing the training time of the proposed model, SupCon-ViT with ViT and other state-of-the-art CNN models, early stopping mechanism with paitance=5}
\begin{center}
\begin{tabular}{lcccc} 
 \toprule
 \textbf{Model} & \textbf{Training Time (in min)}
\\
\midrule
SupCon-ViT & 284 + 25
\\
ViT & 155
\\
ResNet-50 & 198
\\
VGG-16 & 170
\\
DenseNet-121 & 89
\\
MobileNet-V3Large & 61
\\
AlexNet & 56
\\
\bottomrule
\end{tabular}
\end{center}
\label{tab:Time}
\end{table}

The Supervised Contrastive loss used is the pretraining stage of the SupCon-ViT model, where it learns general features and provides an efficient learning framework by maximizing agreement between positive pairs and minimizing agreement between negative pairs. 
This facilitates enhanced feature discrimination, leading to faster convergence and improved classification performance. This efficient learning mechanism allows the SupCon-ViT model to achieve comparable performance with fewer training iterations. The pre-trained feature representation is also used for training the classifier in the classifier stage. This accelerates training by providing a good initialization for the model parameters, leading to faster convergence and better results during classification.


\subsection{Data Augmentation and Hyper-parameters}

In our study, we experimented with various hyperparameter combinations and data augmentations on the IDC dataset to train our model. The optimal learning rate was determined by monitoring the loss during the initial iterations, which allowed us to identify a range of values that consistently led to stable and gradual decreases in loss. The chosen learning rate demonstrated its effectiveness in guiding the model's training by rapidly converging toward lower loss values. For the training of stage 1 and stage 2 of the SupCon-ViT models, we selected learning rates of $5\times10^{-5}$ and 0.01, respectively. 

Previous studies \cite{khosla2020supervised} have indicated that smaller temperature values benefit training more than higher ones, although extremely low temperatures can be challenging due to numerical instability. However, in our case of fine-tuning the model, we observed that lower temperature values could be used effectively. Hence, we aimed to optimize the training procedure and enhance the model's ability to accurately categorize IDC lesions, specifically focusing on achieving a better F1-score. To determine the best data augmentation technique, we evaluated models trained with different augmentations and reported their validation F1-scores in Table \ref{tab:Aug}. Subsequently, we selected the most effective augmentation method and further explored different temperature parameters $\tau$ to refine the model's performance.


\begin{table*}[h]
\caption{Finding best data augmentation for SupCon-ViT model based on validation performance, temperature=0.1}
\begin{center}
\begin{tabular}{lc}   
 \toprule
 \textbf{Data Augmentations} & \textbf{F1-score}
\\
\midrule
AutoAugment \cite{cubuk2019autoaugment} & 0.8163
\\
RandAugment \cite{cubuk2020randaugment} & 0.8132
\\
\textbf{Random Vertical Flip and Random Horizontal Flip} & \textbf{0.8181}
\\
Random Vertical Flip, Random Horizontal Flip, Gray Scale and Color Jitter & 0.8065
\\
 \bottomrule
\end{tabular}
\end{center}
\label{tab:Aug}
\end{table*}

\begin{figure}[h!]
    \centering
    {\includegraphics[scale=0.5]{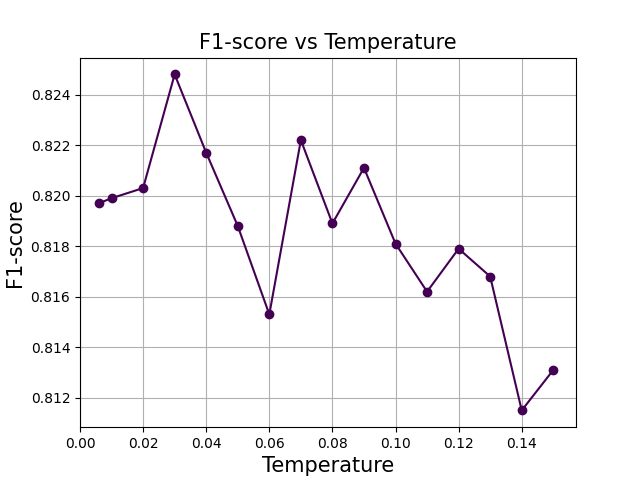}}
    \caption{Finding best temperature for SupCon-ViT model based on validation performance, using best augmentation}
    \label{fig:Temp}
\end{figure}


\section{Conclusion}

In this study, we propose a novel approach for classifying IDC from histopathological images into positive (benign) and negative (malignant) categories. Our method utilizes a pre-trained ViT architecture in conjunction with a supervised contrastive
loss framework. The supervised contrastive loss is employed to enhance the pre-trained ViT model’s discriminative power in capturing the distinctive characteristics of IDC samples and encourages the model to learn representations that are well-separated for positive and negative IDC instances. 
The ViT architecture serves as the foundation for our proposed approach. Capitalizing on the pre-trained ViT enables the model to effectively process and comprehend the spatial information contained in histopathological images. By utilizing self-attention mechanisms, the ViT architecture can capture long-range dependencies and contextual relationships within the image. This empowers the model to learn discriminative features that are relevant to IDC classification.
By combining the strengths of the supervised contrastive loss and the pre-trained ViT architecture, our approach takes advantage of both techniques. Hence, SupCon ViT is proficient at capturing the essential features that differentiate benign and malignant IDC cases which results in improved accuracy in categorizing IDC lesions, thereby providing valuable insights for diagnostic and treatment decisions.

\section{Future Work}

Integrating SupCon ViT with clinical data and gene expression enables a multimodal fusion that incorporates information from different data modalities for a comprehensive and meticulous understanding of breast cancer. Clinical data offers essential details, such as medical history, that can help the model better comprehend and interpret histopathological images. By using histopathological images, clinical data, and gene expression profiles, it would be possible to capture correlations, and interactions between visual features, patient characteristics, and molecular signatures and obtain important cues. Clinical and genomic data integration with SupCon ViTs offers novel opportunities for improving model interpretability and explainability. It would be feasible to pinpoint crucial visual characteristics, clinical elements, or gene expressions that support the model's predictions by examining the learned representations and attention patterns of the model. This could help with early diagnosis, determining a course of therapy, and identifying patients who would gain benefits from certain interventions.



\bibliographystyle{IEEEtran}
\bibliography{main}

\end{document}